%% file: main_arXiv.tex
\pdfoutput=1
\documentclass[11pt]{article}
\usepackage{times}
\usepackage{latexsym}
\usepackage{dblfloatfix}
\usepackage[T1]{fontenc}

\usepackage[utf8]{inputenc}
\usepackage{xurl}
\usepackage{microtype}

\usepackage{inconsolata}
\usepackage{mdframed}
\usepackage{titlecaps}
\usepackage{listings}

\input{math_commands.tex}

\input{config}
\usepackage{acl}

\title{Position Paper: Agent AI Towards a Holistic Intelligence}
\author{
\AND
Qiuyuan Huang$^{\star\complement\blacktriangleright}$,
\And Naoki Wake$^{\star\Re\blacktriangleright\lozenge}$,
\And Bidipta Sarkar$^{\mathsection\dagger}$,
\And Zane Durante$^{\mathsection\dagger}$,
\AND Ran Gong$^{\natural\dagger}$,
\And Rohan Taori$^{\mathsection\dagger}$, 
\And Yusuke Noda$^{\Game}$,
\And Demetri Terzopoulos$^{\natural}$, 
\AND Noboru Kuno$^{\sphericalangle}$,
\And Ade Famoti$^{\sphericalangle}$,
\And Ashley Llorens$^{\sphericalangle}$,
\And John Langford$^{\digamma}$,
\AND Hoi Vo$^{\Game\ddagger}$, 
\And Li Fei-Fei$^{\mathsection\ddagger}$,
\And Katsu Ikeuchi$^{\Re\ddagger}$, 
\And Jianfeng Gao$^{\complement\ddagger}$ 
\AND
  \normalfont$^{\complement}${Microsoft Research Core, Redmond};
  \normalfont$^{\Re}${Microsoft Applied Robotics Research, Redmond}; \\ 
  \normalfont$^{\mathsection}${Stanford University};
  \normalfont$^{\natural}${University of California, Los Angeles}; \\
  \normalfont$^{\Game}${Microsoft Gaming US};
  \normalfont$^{\sphericalangle}${MSR Accelerator};
  \normalfont$^{\digamma}${MSR AI Frontiers, Newyork}
}

\begin{document}

\twocolumn[{%we
\renewcommand\twocolumn[1][]{#1}%
\maketitle
\begin{center}
    \centering
    \vspace{2cm}
    \captionsetup{type=figure}
    \includegraphics[width=0.99\linewidth]{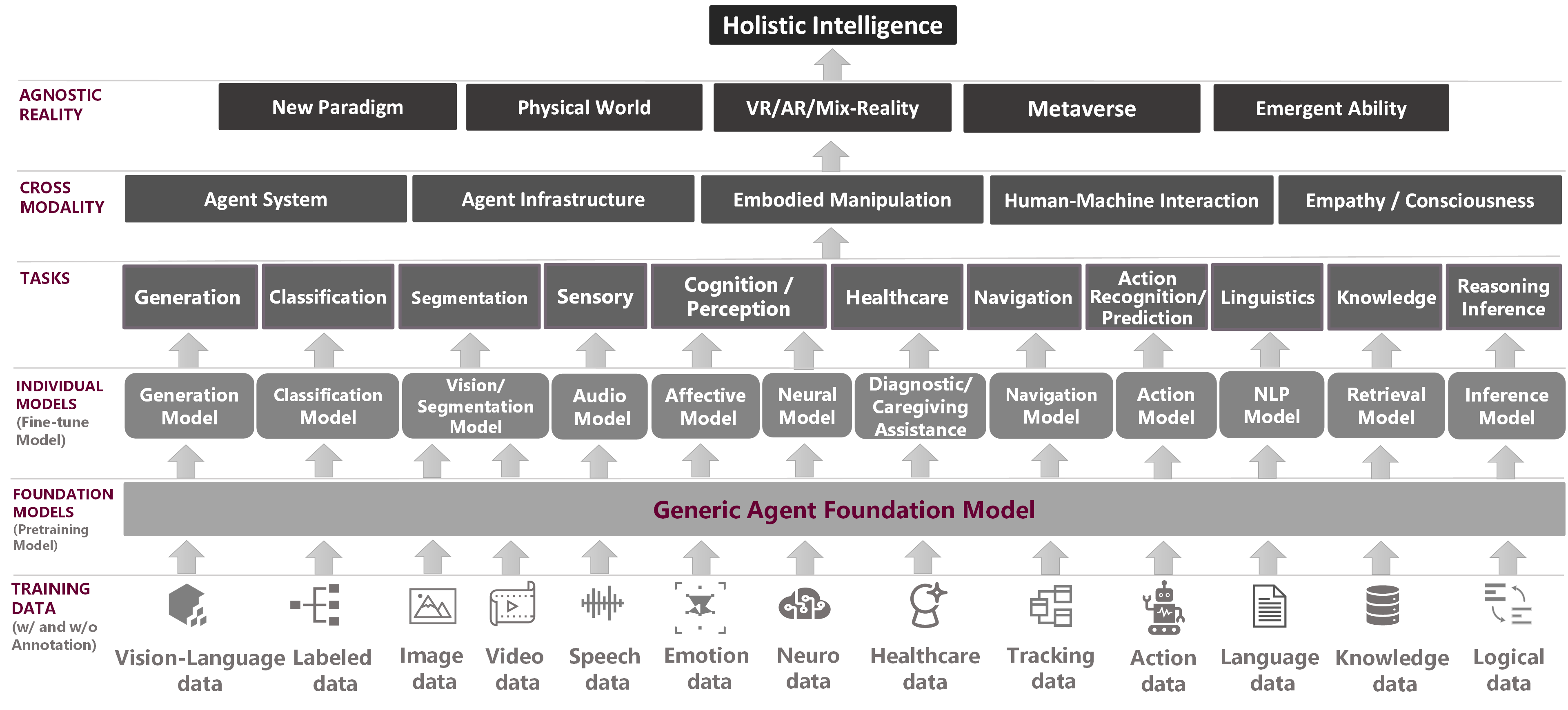}
     \vspace{3mm}
    \captionof{figure}{
    Overview of an Agent AI system. This system is applicable across multiple domains and provides a foundation model for interactive manipulation and embodied operations. Agent AI operates in both physical and virtual worlds by leveraging cross-modal data that is acquired through interactions between diverse environments. 
    Agent AI offers a promising approach to
    unify a broad range of applications and capabilities within infrastructure and system. Furthermore, it is emerging as a promising pathway towards %Artificial General 
    Holistic Intelligence (HI).}
    \vspace{10mm}
    \label{fig:agentai}
\end{center}%
}]

\begin{abstract}

Recent advancements in large foundation models have remarkably enhanced our understanding of sensory information in open-world environments. In leveraging the power of foundation models, it is crucial for AI research to pivot away from excessive reductionism and toward an emphasis on systems that function as cohesive wholes. Specifically, we emphasize developing Agent AI---an embodied system that integrates large foundation models into agent actions. The emerging field of Agent AI spans a wide range of existing embodied and agent-based multimodal interactions, including robotics, gaming, and healthcare systems, etc. In this paper, we propose a novel large action model to achieve embodied intelligent behavior, the Agent Foundation Model. On top of this idea, we discuss how agent AI exhibits remarkable capabilities across a variety of domains and tasks, challenging our understanding of learning and cognition. Furthermore, we discuss the potential of Agent AI from an interdisciplinary perspective, underscoring AI cognition and consciousness within scientific discourse. We believe that those discussions serve as a basis for future research directions and encourage broader societal engagement.
\end{abstract}

\let\thefootnote\relax\footnotetext{$^{\star}$Equal Contribution. $^{\blacktriangleright}$Project Lead. $^{\ddagger}$Equal Advisor.  $^{\lozenge}$Corresponding Author. 
$^{\dagger}$Work done while interning or researching part-time at Microsoft Research, Redmond.}

\clearpage
%%%%%%%%%%%%

\input{sections/0_intro}
\input{sections/1_agentAI_paradigm}
\input{sections/2_agentAI_categorization.tex}
\input{sections/3_agentAI_applications.tex}
\input{sections/4_challenge.tex}

\input{sections/5_influences.tex}

\input{sections/6_conclusion.tex}

\input{sections/7_broade_impact}

%\input{icml2024/sections/6_discussion}
%\input{sections/8_appendix}
%\input{icml2024/sections/submission_instructions} % remove for final version, just here so everyone can read.

%\bibliographystyle{acl_natbib}
%\bibliography{custom}
\bibliography{main}

\clearpage

\appendix

%%%%%%%%%%%%%%%%%%%%%%%%%%%%%%%%%%%%%%%%%

% \vbox{
%     \centering
%     \large 
%     \textbf{Appendices for \\  
%     MindAgent: Emergent Gaming Interaction}
% }

\twocolumn[
  \begin{@twocolumnfalse}
    \begin{center}
      \Large \textbf{Appendices for \\ Agent AI Towards a Holistic Intelligence}
    \end{center}
    \vspace{5mm} % Adjust the vertical space as needed
  \end{@twocolumnfalse}
]
%%%%%%%%%%%%%%%%%%%%%%%%%%%%%%%%%%%%%%%%%
% \vskip 0.18in
% \vskip -\parskip
% \hrule height 1pt
%%%%%%%%%%%%%%%%%%%%%%%%%%%%%%%%%%%%%%%%

%%%%%%%%%%%%%%%%%%Appendix
%%%%%%%%%%%%
\section{Intention Information and Manipulation for Embodied Action}
\label{app:Intention}
Language Conditioned internet action instruction entails the ability of a robotic system to interpret and execute tasks based on language instructions. This aspect is particularly crucial for creating intuitive and user-friendly interfaces for human-robot interaction. Through natural language commands, users can specify goals and tasks to robots in a manner similar to human-human communication~\cite{wang2019reinforced}, thereby lowering the barrier to operating robotic systems. In a practical scenario, for instance, a user could instruct a service robot to ``pick up the red apple from the table,'' and the robot would parse this instruction, identify the referred object and execute the task of picking it up~\cite{wake_chatgpt}. The core challenge lies in developing robust natural language processing and understanding algorithms that can accurately interpret a wide array of instructions, ranging from direct commands to more abstract directives, and enable the robot to convert these instructions into actionable tasks. Furthermore, ensuring that robots can generalize these instructions across diverse tasks and environments is critical for enhancing their versatility and utility in real-world applications.
The use of language input to guide robot's task planning has gained attention in the context of a robot framework called Task and Motion Planning~\cite{garrett2021integrated}. 

In addition, ~\cite{durante2024foundation}learn about the intricate challenges of large action models for embodied systems e.g., robotic. It begin with a low-level  action manipulation foundational models, it explore solutions to issues such as action resignation, adaptability to dynamic environments, and the efficient management of high-dimensional action spaces. When transfer to next phase, we implement and refine algorithms, ensuring scalability and effectiveness in simulations on our server. Develop and implement foundational algorithms for large action models, emphasizing efficiency and scalability. Focus on addressing issues related to pre-training, fine-tuning, and model optimization. Conduct initial simulations on Azure to validate algorithmic concepts. The ultimate objective in the third phase, is to optimize and validate these algorithms in real-world scenarios, exploring diverse applications and contributing to the evolution of large action models for General Purpose Robotics. Optimize and validate the algorithms in real-world scenarios with large action models on Phoenix. Explore applications in diverse domains, ensuring robustness and scalability. Refine algorithms based on real-world evaluation feedback and scale for broader cloud deployment in the embodied system.

%%%%%%%%%%%%%

\section{Agent for Cross-modality and Mix-reality}
\subsection{Agents for Cross-modal Understanding}
\label{app:Crossmodal}
Multi-modal understanding is a significant challenge for creating generalist AI agents due to the lack of large-scale datasets that contain vision, language, and agent behavior.  More generally, training data for AI agents is often modality specific.  This results in most modern multi-modal systems using a combination of frozen submodules.  Some notable examples are Flamingo \cite{alayrac2022flamingo}, BLIP-2 \cite{li2023blip}, VLC \cite{gui2022vlc} and ArK \cite{huang2023ark}, all of which utilize a frozen LLM and frozen visual encoder.  These submodules are trained individually on separate datasets, and then adaptation layers are trained to encode the visual encoder into the LLM embedding space.  In order to make further progress for cross-modal understanding for AI agents, it is likely that the strategy of using frozen LLMs and visual encoders will need to change.  Indeed, RT-2, a recent visual-language model that is capable of taking actions within the domain of robotics showed significantly improved performance when jointly tuning the visual encoder and LLM for robotics and visual-language tasks \cite{brohan2023rt}.

\subsection{Agents for Cross-domain Understanding}
\label{app:Crossdomain}
A key challenge for creating generalist agents is the distinctive visual appearance and disparate action spaces across different domains.  Humans possess the capability to interpret images and videos from various sources, including the real world, video games, and specialized domains such as robotics and healthcare ~\cite{durante2024agent}, once they become familiar with the specific details of these areas.  However, existing LLMs and VLMs often demonstrate significant differences between the data they were trained on and the varied domains in which they are applied. And notably, training agent models to predict specific actions presents a considerable challenge when trying to develop a single policy that can effectively learn multiple control systems across domains~\cite{huang2023ark}.  Generally, the approach most modern works take when applying systems within specific domains is to start from a pretrained foundation model and then finetune a separate model for each specific domain.  This fails to capture any commonalities between domains and results in a smaller total set of data used for training instead of leveraging each domain's data.

%Large pretrained models such as CLIP, ViLD and PaLI which enable few-shot and zero-shot performance on novel tasks.
\subsection{Interactive agent for cross-modality and cross-reality}
\label{app:ModalReality}
Developing AI agents that can successfully understand and perform tasks across different realities is an on-going challenge that has seen some recent success for image and scene generation \cite{huang2023ark}.  In particular, it is challenging for agents to simultaneously understand real-world and virtual reality environments due to their visual dissimilarities and separate environment physics.  Within the context of cross-reality, Sim to Real transfer is a particularly important problem when using simulation-trained policies for real-world data, which we discuss in the next section.

%%%%%%%%%%%%%%%

%%%%%%%%%%%%
\section{Bias}
\label{app:Bias}
AI agents based on LLMs or LMMs (large multimodal models) have biases due to several factors inherent in their design and training process. When designing these AI agents, we must be mindful of being inclusive and aware of the needs of all end users and stakeholders. In the context of AI agents, \textit{inclusivity} refers to the measures and principles employed to ensure that the agent's responses and interactions are inclusive, respectful, and sensitive to a wide range of users from diverse backgrounds.
Despite these measures, AI agents still exhibit biases. Ongoing efforts in agent AI research and development are focused on further reducing these biases and enhancing the inclusivity and fairness of agent AI systems. Despite these measures, AI agents still exhibit biases. Ongoing efforts in agent AI research and development are focused on further reducing these biases and enhancing the inclusivity and fairness of agent AI systems. Despite these efforts, it's important to be aware of the potential for biases in responses and to interpret them with critical thinking. Continuous improvements in AI agent technology and ethical practices aim to reduce these biases over time. One of the overarching goals for inclusivity in agent AI is to create an agent that is respectful and accessible to all users, regardless of their background or identity.

%%%%%%%%%%%
\section{Hallucinations}
\label{app:Hallucinations}
Agents that generate text are often prone to hallucinations, which are instances where the generated text is nonsensical or unfaithful to the provided source content \cite{raunak2021curious,maynez-etal-2020-faithfulness}. Hallucinations can be split into two categories, \textit{intrinsic} and \textit{extrinsic} \cite{ji2023survey}.  Intrinsic hallucinations are hallucinations that are contradictory to the source material, whereas extrinsic hallucinations are when the generated text contains additional information that was not originally included in the source material.

Some promising routes for reducing the rate of hallucination in language generation involve using retrieval-augmented generation \cite{lewis2020retrieval,shuster2021retrieval} or other methods for grounding natural language outputs via external knowledge retrieval \cite{dziri2021neural,peng2023check}.  Generally, these methods seek to augment language generation by retrieving additional source material and by providing mechanisms to check for contradictions between the generated response and the source material.

Within the context of multi-modal agent systems,  have multimodality been shown to hallucinate as well \cite{zhou2023analyzing}.  One common cause of hallucination for vision-based language-generation is due to the over-reliance on co-occurrence of objects and visual cues in the training data \cite{rohrbach2018object}. AI agents that exclusively rely upon pretrained large foundation models and use limited environment-specific finetuning can be particularly vulnerable to hallucinations since they rely upon the internal knowledge-base of the pretrained models for generating actions and may not accurately understand the dynamics of the world state in which they are deployed.

\end{document}

%% file: math_commands.tex
\usepackage{amsmath,amsfonts,bm}

\def\eqref#1{equation~\ref{#1}}
\def\1{\bm{1}}

\DeclareMathAlphabet{\mathsfit}{\encodingdefault}{\sfdefault}{m}{sl}
\SetMathAlphabet{\mathsfit}{bold}{\encodingdefault}{\sfdefault}{bx}{n}

%% file: config.tex
\definecolor{Gray}{gray}{0.9}
\definecolor{mygreen}{rgb}{0.0, 0.5, 0.0}
\definecolor{myred}{rgb}{0.8, 0.25, 0.33}
\definecolor{myblue}{rgb}{0.19, 0.55, 0.91}
\definecolor{uclablue}{rgb}{0.15, 0.45, 0.68}
\definecolor{boxgreen}{rgb}{0.02, 0.66, 0.02}
\definecolor{boxred}{rgb}{0.66, 0.1, 0.1}
\definecolor{boxblue}{rgb}{0.01, 0.01, 0.73}
\usepackage{url}
\usepackage{etoc}
\usepackage{amsmath}
\usepackage{booktabs}
\usepackage{algorithm}
\usepackage{algorithmic}
\usepackage{lipsum}
\usepackage{pifont}
\usepackage{wrapfig}
\usepackage{colortbl}
\usepackage{acronym}
\usepackage{multicol}
\usepackage{multirow}
\usepackage{makecell}
\usepackage{enumitem}
\usepackage{tabularx,colortbl,array}
\usepackage[skip=3pt,font=small]{subcaption}
\usepackage[skip=3pt,font=small]{caption}
\usepackage[breaklinks,pagebackref,citecolor=uclablue,colorlinks=true,linkcolor=black]{hyperref}
\usepackage{xspace}
\usepackage{mdframed}

\usepackage{mathtools}
\usepackage{amssymb}
\usepackage{amsthm}

\renewcommand{\paragraph}[1]{\noindent\textbf{#1.}}
\makeatletter
\DeclareRobustCommand\onedot{\futurelet\@let@token\@onedot}
\def\@onedot{\ifx\@let@token.\else.\null\fi\xspace}

\makeatother

\acrodef{llms}[LLMs]{Large Language Models}
\acrodef{mlms}[MLMs]{Multimodal Language Models}

\usepackage{xcolor}

\definecolor{mygray}{gray}{0.4}

%% file: sections/0_intro.tex
\section{Introduction}
\label{sec:intro}

\begin{figure*}[t]
    \centering  \includegraphics[width=0.95\linewidth]{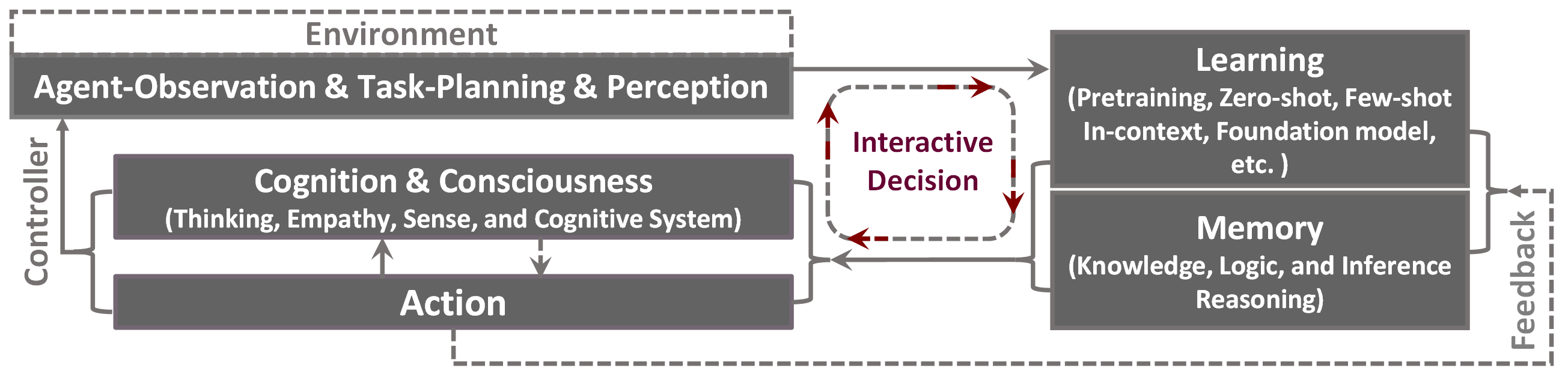}
    \vspace{2mm}
    \caption{An Agent AI paradigm for supporting embodied multi-modal generalist agent systems. There are five main modules as shown: (1) Agent in Environment and Perception with task-planning and observation, (2) Agent Learning, (3) Memory, (4) Action, and (5) Cognition and Consciousness. We believe that the cohesive integration of these components facilitates the development of a holistic intelligence. A key distinction of our approach from some prior interactive strategies is that, after training, the agent's actions will directly influence task planning without the need for receiving feedback from the environment to plan its subsequent actions as the previous interacive paradigm.}
    \label{fig:agentpara}
\end{figure*}

%%%%%%%%%%%%%The following two paragraphs wroten by Katsu, I copied here again.  --Qiuyuan

Artificial Intelligence (AI) was historically defined at the 1956 Dartmouth Conference as artificial life forms capable of collecting information from their environment and taking effective actions within it. Minsky's group at MIT developed a robotic system in 1970, known as the ``Copy Demo,'' that observed ``blocks world'' scenes and successfully reconstructed the observed polyhedral block structures~\cite{winston1972mit}. The system, comprising observation, planning, and manipulation modules, demonstrated that each of these subproblems was highly challenging and necessitated further research. Consequently, the field of AI fragmented into specialized subfields. While these subfields have made significant progress independently, this over-reductionism has blurred the overarching goals of AI research.

To advance beyond the current state towards more sophisticated AI, we emphasize the importance of embracing the holistic philosophy of Aristotle, which underscores the integration of components to surpass the sum of its parts. Recent advancements in Large Language Models (LLMs) and Visual Language Models (VLMs) have shown great potential in recognizing language and images in an open-world context~\cite{gpt4}. For example, the advanced semantic processing of LLMs has been utilized to decompose human instructions into high-level tasks for robots~\cite{wake2023gpt4vision, wake2023gpt}. However, these existing multimodal foundation models, even for GPT-4V(ision), still face a challenge in  achieving fine-grained manipulation that necessitates action prediction. Therefore, a new embodied Agent Foundation Model was proposed~\cite{durante2024foundation} which integrates language proficiency, visual cognition, context memory, intuitive reasoning, and can predict the embodied actions with adaptability. This is the first study that pretrains a foundation model for the development of general-purpose AI agents by using embodied data collected from robotics, gaming, and healthcare tasks. %The work is a significant stepping stone towards holistic intelligence.  

%%%%%%%%%%%%%%%
%Importantly, 
An embodied agent is conceptualized as an interactive system that communicates with humans and interacts with environments through its perceptual capabilities, employing actions aligning with human intents. This is the reason why we consider the advance of large embodied foundation models as a significant contribution to Agent AI, enabling systems to parse and infer human intent from various domain information, actions, natural-language instructions and multimodal contexts. Moreover, actively leveraging action-based large foundation models makes our approach unique for developing integrated AI systems. 
%[Jianfeng: I am not sure about the claim of our uniqueness based on recent survey papers -- many people are build action foundation models for agent ai. it is fair to say that we are making substantial contribute to the research.]
%【Qiuyuan: At firstly, the paragraph not only talk about our agent-fondation-pretraining model, this is about the genaral fondation model (also cover Katsu's group zero shot project). Regarding real Action foundation model (real embodied pretraining level), as I know, by now, only google RT1/2 and ours. Not too many. others, as I know, 1) only for the high-level action instruction generation, not low-level embodied action prediction, it can't be called as embodied foundation pretraining. 2) Even for google RT1/RT2, only work on one robotics tasks, if only have one task to do the pretraining, we can't be called as real "pretraining".]
%in the history of AI research aimed at integrated systems.
%%%%%%%%%%%%%%%%

Building upon the Agent AI framework, we believe that the AI community will steadily accumulate insights and knowledge essential for transitioning from AI models used for passive, structured tasks to those capable of dynamic, interactive roles in complex environments. This is a critical step towards the development of Artificial General Intelligence (AGI) (Fig.~\ref{fig:agentai}).
In this paper, we analyze a new architecture for Agent AI systems, alongside a review of recent literature in Agent AI domains including robotics, gaming, and healthcare. Furthermore, we explore the cognitive aspects of Agent AI and introduce research areas impacted by Agent AI to engage a broader community of researchers and actively promote its development. Finally, we discuss future research directions, including the ethical challenges that need to be addressed. Through these discussions, we aim to illustrate how the development of these technologies is bringing AI agents closer to AGI, holistic intelligence.

%% file: sections/1_agentAI_paradigm.tex
%\section{Agent AI Paradigm}
\section{Agent AI Paradigm}
\label{sec:paradigm}
%\subsection{Agent AI Paradigm}

%Agent AI new paradigm represents a change in thinking in embodied intelligence, emphasizing the importance of complex dynamics, and an integrated approach to interactive intelligence. This approach is motivated by the belief that true intelligence arises from the intricate interplay between learning, memory, action, observation, planning, perception, and cognition in a interactive decision with consciousness.

\subsection{Agent AI fundamentals} 
We define Agent AI as an \textit{intelligent agent capable of autonomously executing appropriate and contextually relevant actions based on sensory input, whether in a physical, virtual, or mixed-reality environment.} Agent AI represents a new paradigm that sheds light on embodied intelligence, emphasizing the importance of an integrated approach for interactive agents in complex dynamics. This approach is motivated by the belief that intelligence arises from the intricate interplay between learning, memory, action, perception, planning, and cognition (Fig.~\ref{fig:agentpara}).

%We can build a neuro-cognitive module that can be deployed onto its embodied robots, and generalizable into other agents through a cloud service. The deployed cognitive on the infinite agent will give the ability to understand and respond to dynamic, real-world situations, making them potentially more versatile and adaptive in complex environments. 

%We define the Agent AI as \textit{``any intelligent agent capable of autonomously taking suitable and seamless action based on sensory input, whether in the physical world or in a virtual or mixed-reality environment representing the physical world.''}. Importantly, an embodied agent is conceptualized as a collaborative system, where it communicates with humans with its vision-language capabilities and employ a set of vast actions based on human needs. In this manner, embodied agents are expected to mitigate cumbersome tasks in virtual reality and physical world. 

%\vspace{-10mm}
\paragraph{Learning} Agent AI can adapt to new environments by acquiring new knowledge and updating its skills. To this end, the agent needs to observe its environment, understand the impact of its actions on that environment, and learn from human demonstrations~\cite{wake2020learning}. For instance, by employing reinforcement learning (RL) techniques or supervised learning from human demonstrations (e.g., imitation learning (IL),  behavior cloning), the agent can progressively improve its behavior.%Additionally, the agent should always be under human supervision for safety. In case it encounters a difficult situation, it should ask the user for help and further instructions.

\paragraph{Memory} Long-term memory enables the Agent to remember specific operations adaptable to the environment or user preference. In contrast, short-term memory pertains to the history of actions taken and perceptions observed during an operation. Short-term memory enables the system to replan and consider next-step actions based on history.

\paragraph{Action} The actions of Agent AI do not necessarily have to be physical actions in the real world. Depending on the definition of the environment, actions may include interactions in virtual reality (VR) environments or speech directed at humans. A suitable action is selected through a cognitive process from learned skills, based on memory. Additionally, real-world operations often cannot be completed in one shot and thus require multi-round interactions between humans or the environment and the agent. This interaction is also orchestrated by a cognitive process and memory (e.g., conversation history).

\paragraph{Perception} Like humans, robust and multimodal perception is crucial for agents to understand their environment. Visual perception is one of the most important abilities, enabling the agent to comprehend the world, e.g., images, videos, gameplay. Similarly, audio perception is crucial for understanding human intent.

\paragraph{Planning} Planning is an important aspect of long-range tasks, such as a robot manipulating objects in an environment for a specific purpose. %navigating a robot in an environment with a specific purpose. 
The planning strategy typically depends on the goal of the task. Goal-oriented planning enables flexible operation that adapts to uncertainties due to any external and internal disturbances.

\paragraph{Cognitive Aspects}
Agent AI focuses not only on the performance of individual components but also on the utility of the system as a whole. Consider a scenario where a robot, right after being unboxed, begins to communicate  with a non-expert user and swiftly adapts to carry out domestic tasks within the user's home setting. Realizing such a system is challenging and requires a mechanism that orchestrates each Agent AI components. This orchestration functionality is referred to as the cognitive aspect of Agent AI.

%We can build a neuro-cognitive module that can be deployed onto its embodied robots, and generalizable into other agents through a cloud service. The deployed cognitive on the infinite agent will give the ability to understand and respond to dynamic, real-world situations, making them potentially more versatile and adaptive in complex environments. 
%we believe that it is crucial to involve diverse researchers in Agent AI to surmount this gap.

\subsection{Agent AI Consciousness} 
Agent AI can go beyond a simple component orchestration and potentially entail a type of ``consciousness.'' In recent challenging attempts to find consciousness in AI based on neuroscientific insights, neuroscientists have discussed \textit{Agency} and \textit{Embodiment} as indicators of consciousness~\cite{butlin2023consciousness}. Agency refers to the capacity to learn from feedback, make decisions to pursue goals, and adapt to conflicting objectives. It indicates a system's characteristic of attempting to achieve goals through interaction with its environment. Embodiment involves understanding and utilizing the relationship between actions and feedback from the environment to affect perception or control. It emphasizes comprehending how one's body and the surrounding environment can be leveraged in cognitive processes.

Our Agent AI predicts optimal actions based on language (i.e., textual instructions), sensory inputs, and action history, fulfilling Agency by generating goal-directed actions. It also learns from the relationship between its actions and environmental outcomes, fulfilling the principle of Embodiment. Thus, we can potentially quantify aspects of Agent AI's consciousness, suggesting its potential across disciplines like neuroscience, biology, physics, biological physics, cognitive science, medical health, and moral philosophy. %In this paper, we argue, that consciousness/cognition of AI is best assessed by drawing on neuroscientific theories of consciousness. We describe prominent theories of this kind and investigate their implications for agent AI.

There are various approaches to developing Agent AI. In Section~\ref{sec:category}, we will introduce a specific example of Agent AI. In Section~\ref{sec:discussion}, we will discuss the main challenges and necessary actions, including ethical concerns in Agent AI research.

%Despite numerous gaps between current technologies and holistic intelligence, the recent advancements in LLMs/VLMs have brought society closer to the idea that such a system is within reach. What steps are required to achieve this ultimate goal? In light of traditional AI philosophy, we believe that successful Agent AI systems require several key components: 

%With our \textbf{Interactive Agent Transformer} serving as an initial step in this direction, the next phase of our research will further align with the new embodied agent Paradigm. To this end, we intend to engage a broad range of experts and practitioners to discuss important research areas, including but not limited to: 
%%%%%%%%%%%%%%%%%%

\section{Agent Foundation Model}
%There is no fixed design method to follow the principle outlined in Section \ref{sec:paradigm}. 
Agent AI systems that interact with the environment, with humans, and amongst other agents.  We consider agent-environment interactions as encompassing a broader scope than embodied agents.  For instance, ambient intelligence systems, which despite not only having a physical embodiment, can be embedded into and interact with their environment. The advancement of agent systems that interact with humans is another area of keen interest for this area. We  strongly believe that multimodal interactions between humans and agents, extending beyond high-level intention instructions, is a promising area of research and future direction for low-level fine-grained actions manipulation with human-agent interactions.  We are also interested indeveloping systems for effective agent to agent communication and efficient collaboration within multi-agent infrastructures and exploring new agent paradigm and agent learning Strategy. 

In this section, we provide an overview of Agent AI system that leverages foundation models with the latest machine-learning technologies. %(Fig.~\ref{fig:Agentfodation}). 
The system is composed of three components: i) Interactive agent transformer, ii) Agent foundation model learning strategy with RL and IL, and iii) self improvement.

\subsection{Agent Transformer}
\begin{figure}[th]
    \centering
    \includegraphics[width=0.99\linewidth]{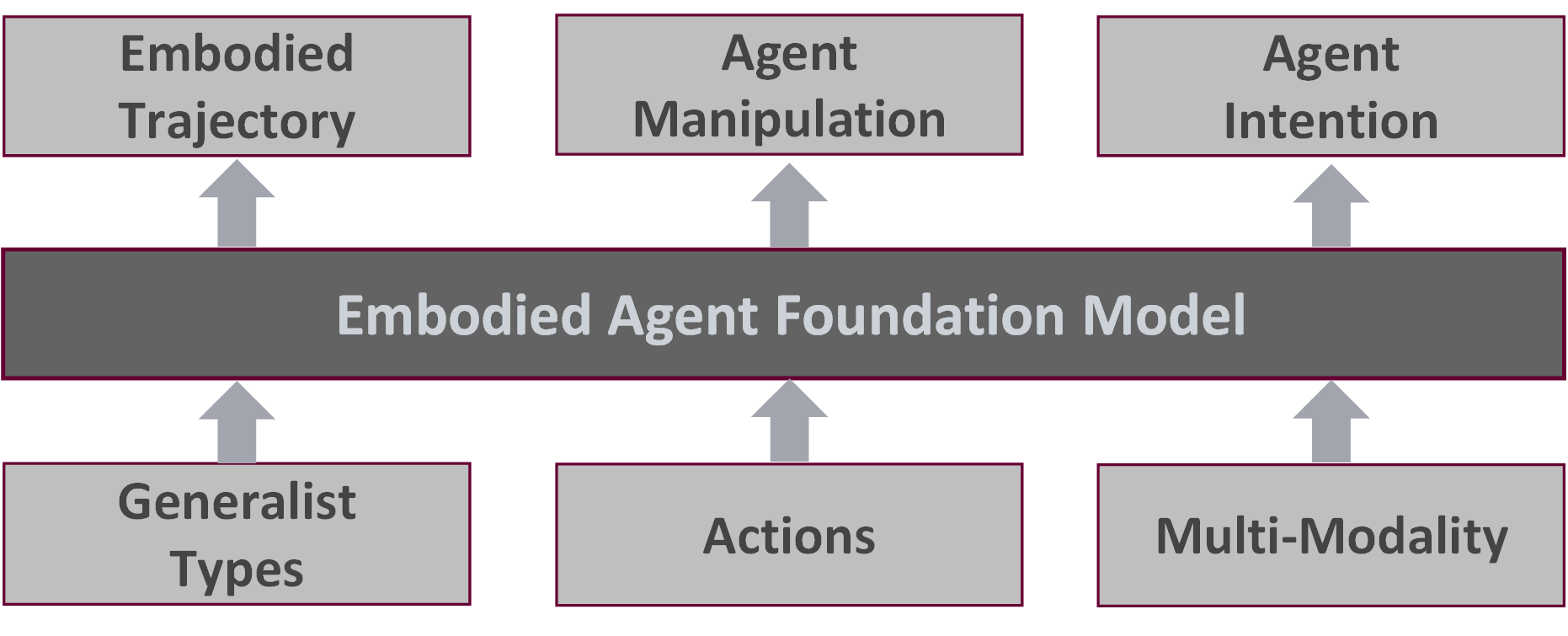}
     \vspace{-2mm}
    \caption{Overview of an interactive agent foundation model framework. The transformer is designed to process multi-modal information that conveys various levels of abstraction. This approach facilitates a comprehensive understanding of the context, thus enhancing coherent actions. Through learning across a variety of task domains and applications.
    }
\vspace{-1mm}
    \label{fig:Agentfodation}
\end{figure}

We analyze a transformer-based multimodal encoder (Fig.~\ref{fig:Agentfodation}) that enables an interactive agent to take actions based on multimodal information. This model is initialized with three pre-trained submodules, namely, the visual module, the agent action module and the language module.

%CLIP ViT-B16, which is used to initialize our visual encoder, and OPT-125M, which is used to initialize our action and language model.

%In our approach, each frame in a video is encoded as visual features. 
To facilitate cross-modal information sharing, %we train a linear projection layer to transform the visual features generated by the visual encoder into visual tokens in the same embedding space of language. 
~\cite{durante2024foundation} foundation model allows the agent to predict actions (or action tokens) to complete the embodied tasks in robot, gaming, and interactive healthcare domains.
%[Jianfeng: I don't understand what you want to say here. Please check.]
%general intention token, and any other embodied types for training the model to perform agent's behaviors in specific domains.% text prompt, agent prompt and a single/a series of video frame.
The model also feed diverse historical data into the transformer model including but not limited to previous low-level fine-grained actions (agent information),  video/images, audio, language, or high-level instruction, as context during pre-training. As a result, for any given time step, it can predict low-level manipulation (action) tokens, general agent types (e.g., TypeChat in gaming), or high-level instructions (e.g., agent intention).
Moreover, the unified transformer can also produce high-level instructions based on text prompts, visual context, and previous actions. This approach allows the model to take into account both the current context and the history of interactions, making it able to respond more accurately to the task at hand.
%[Jianfeng: these two paragraphs are repetitive, and can be merged into one.]

\subsection{Agent Learning Strategy}
\paragraph{Reinforcement Learning (RL)} To learn the optimal relationship between states and actions based on rewards (or
penalties) received as a result of its actions, we can use reinforcement learning. RL is a highly scalable framework that has been applied to numerous applications, including robotics. For many applications, it is challenging or costly to collect human demonstrations, such as learning policies in automatically generated virtual environments. RL is particularly effective in these scenarios, exemplified by the actor-critic algorithm PPO~\cite{schulman2017proximal}. Additionally, RL technology can be applied to model human-AI interactions, which is a crucial aspect of interactive Agent AI. For instance, agents can be trained via RL from human feedback (RLHF) ~\cite{ouyang2022training}, allowing humans to choose desired responses without hand-engineering rewards.

\paragraph{Imitation Learning (IL)}
IL seeks to leverage demonstration data to mimic the actions of human experts. For example, in robotics, one of the major frameworks based on IL is Behavioral Cloning (BC). BC is an approach where a robot is trained to mimic the actions of an human expert by directly copying them. In this approach, the expert's actions in performing specific tasks are recorded, and the robot is trained to replicate these actions in similar situations. Recent BC-based methods often incorporate technologies from LLM/VLMs, enabling more advanced end-to-end models. For example, Brohan et al. proposed RT-1~\cite{brohan2022rt} and RT-2~\cite{brohan2023rt}, transformer-based models that output an action sequence for a robot's base and arm, taking a series of images and language as input. These models are reported to show high generalization performance as the result of training on a large amounts of demonstration data.

\paragraph{Traditional RGB}
Learning intelligent agent behavior leveraging image inputs has been of interest for many years \cite{mnih2015human}.  The inherent challenge of using RGB input is the curse of dimensionality. To solve this problem, researchers either use more data \cite{jang2022bc, ha2023scaling} or introduce inductive biases into the model design to improve sample efficiency.  In particular, authors incorporate 3D structures into the model architecture for manipulations \cite{zeng2021transporter, shridhar2023perceiver, goyal2023rvt, james2022q}. For robot navigation, authors \cite{chaplot2020object, chaplot2020neural} leverage maps as a representation. Maps can either be learned from a neural network aggregating all previous RGB inputs or through 3D reconstruction methods such as Neural Radiance Fields \cite{rosinol2022nerf}.
%To obtain more data, researchers synthesize synthetic data using graphics simulators \cite{mu2021maniskill, gong2023arnold}, and try to close the sim2real gap \cite{tobin2017domain, sadeghi2016cad2rl, peng2018sim}.  Recently, there has been some collective effort to curate large-scale dataset that aims to resolve the data scarcity problem \cite{padalkar2023open, brohan2023rt}. On the other hand, to improve sample complexity, data augmentation techniques have been extensively studied as well \cite{zeng2021transporter, rao2020rl, haarnoja2023learning, lifshitz2023steve}.
%We can leverage existing learning techniques to facilitate the learning of agent policies. For example, IL is an optimal approach when the human demonstration data is available. Recent transformer-based models have been reported to show high generalization performance as a result of training on a large amount of training data~\cite{brohan2022rt,brohan2023rt}.
 %We can use the actor-critic algorithm PPO~\cite{schulman2017proximal} to update the parameters of the agent using its own version.

%\paragraph{Generation Agent with Imitation Learning (IL)}

\subsection{Optimization in the Agent System}
The optimization of agent systems can be divided into spatial and temporal aspects. Spatial optimization considers how agents operate within a physical space to execute tasks. This includes inter-robot coordination, resource allocation, and keeping an organized space. In order to effectively optimize agent AI systems, especially systems with large numbers of agents acting in parallel, previous works have focused on using large batch reinforcement learning \cite{shacklett23madrona}. Since datasets of multi-agent interactions for specific tasks are rare, self-play reinforcement learning enables a team of agents to improve over time. However, this may also lead to very brittle agents that can only work under self-play and not with humans or other independent agents since they over-fit to the self-play training paradigm. To address this issue, we can instead discover a diverse set of conventions~\cite{cui2023adversarial, sarkar2023diverse}, and train an agent that is aware of a wide range of conventions. Foundation models can further help to establish conventions with humans or other independent agents, enabling smooth coordination with new agents~\cite{gong2023mindagent}.

Temporal optimization, on the other hand, focuses on how agents execute tasks over time. This encompasses task scheduling, sequencing, and timeline efficiency. For instance, optimizing the trajectory of a robot's arm is an example of efficiently optimizing movement between consecutive tasks~\cite{zhou2023generalizable}. At the level of task scheduling, methods like LLM-DP~\cite{dagan2023dynamic} and ReAct~\cite{yao2023react} have been proposed to solve efficient task planning by incorporating environmental factors interactively. 

%%%%%%%%%%%%%%%%
%\subsection{Iterative Improvement of Agent AI}
%\vspace{-4mm}
\subsection{Self Improvement for Transformers}
Currently, foundation model based AI agents have the capacity to learn from multiple different data sources, which allow for more flexible sources for data for training.  Two key consequences of this are that (1) user and human-based interaction data can be used to further refine and improve the agent and (2) existing foundation models and model artifacts can be used to generate training data.  We discuss each of these in more detail in the following sections, but we note that since current AI Agents are largely tied to existing pretrained foundation models, they generally do not learn from continuous interaction with their environments.  We think this is an exciting future direction, and initial work by Bousmalis et al. has shown that self-improving agents for robotic control are able to continuous learn and improve through environmental interactions without supervision \cite{durante2024foundation,bousmalis2023robocat}.

%Once trained on large volumes of text and images, the foundation model can generate images based on the data it has been trained on. Agent AI utilizes its generative capabilities within its information processing framework. For instance, a VLM can reconstruct a specific image from an input instruction. This image is then further refined by referring to the Agent memory, leading to an action based on this enhanced memory. The system then uses this output to reflect and generate a new prompt, serving as input for the VLM in the subsequent step. This iterative loop facilitates the self-adaptation of the Agent AI in its environment, enabling continuous learning and refinement of its responses.

Furthermore, the iterative learning process can leverage human feedback~\cite{gong2023mindagent}. For example, in the context of robot teaching, Agent AI understands what it needs to do from multimodal instructions provided by humans~\cite{wake2020learning}. Based on these instructions, it generates images or scenes and makes them operable in a virtual world. This process is repeated by utilizing user feedback, allowing Agent AI to gradually improve and adapt itself to the environment.% (e.g., robot grasping skill~\cite{saito2022task}). %More information about generating the prompt to run the simulation robot can be referenced in ~\cite{wake2023gpt4vision} and ~\cite{wake2023gpt}.
\vspace{2mm}

%% file: sections/2_agentAI_categorization.tex
%\vspace{-3mm}
\section{Agent AI Categorization}
\label{sec:category}
%[Jianfeng: Need to clarify how we group AI Agents into different categories, e.g., based on eniroment (physical, digital, mixed), or tasks (knowledge/emotional reasoning), or implementation (neural approach, symbolic approach). ]
Agent AI aims to develop agents that can adeptly navigate and interact with a changing world. These agents are designed to learn and solve complex tasks through direct engagement with their environment. The field has been propelled forward by significant advancements in the development of general-purpose foundation models, leading to superhuman achievements in various AI domains previously deemed challenging.  These developments have significantly boosted the capabilities of embodied AI. Researchers are now rapidly advancing towards creating intelligent agents that can perceive their surroundings, engage in natural language dialogue, understand and respond to auditory inputs, navigate and manipulate their environment to achieve objectives, and reason about the long-term outcomes of their actions.  We are interested in particular with submissions that focus on the multimodal aspects of embodied AI systems and develop novel methods for synthesizing meaningful agent outputs from multi-sensory inputs. 

\subsection{Embodied Agent Categorization}

\begin{figure}[th]
    \centering
\includegraphics[width=0.98\linewidth]{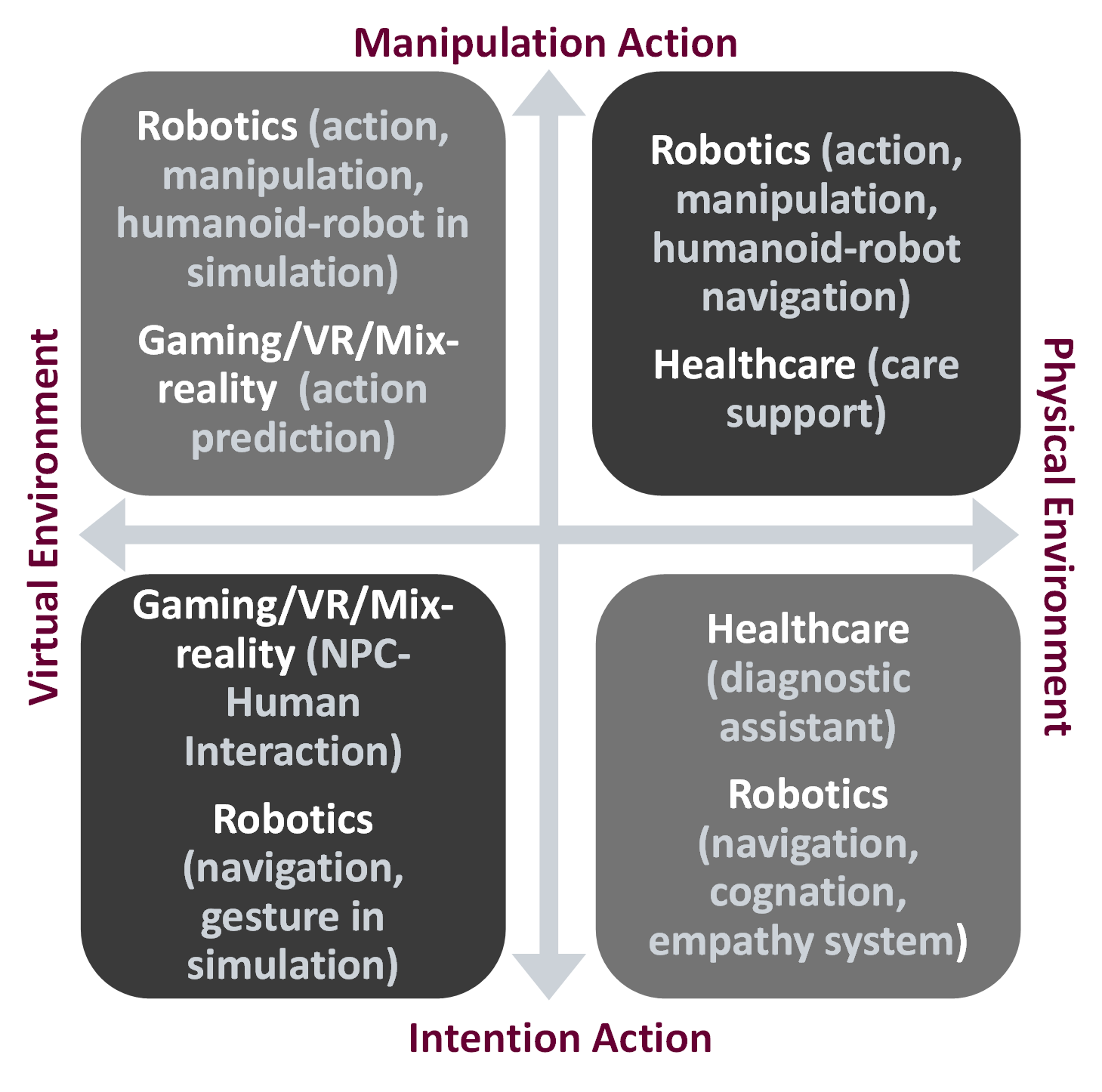}
     %\vspace{2mm}
    \caption{Overview of the two axes for agents spaces. Embodied Agent AI is classified according to the extent to which it involves low-level fine action manipulations, which we refer to as ``manipulation actions'' (e.g., action prediction) in an environment, whether real or virtual. In contrast, an agent's actions may primarily aim at high-level information transmission for a robot or human's intent instruction, which we refer to as ``intention action'' (e.g., general task planning). An agent's environment can be broadly categorized based on whether it is the physical world or a virtual one. According to this, we divide embodied and interactive Agent into main four categories.
    }
    \vspace{-3mm}
    \label{fig:agentaxes}
\end{figure}

Agent AI refers to AI systems that integrate Large foundation models. Consequently, a number of recent AI systems that are based on LLM/VLMs can be associated with Agent AI subcategories. Specifically, we categorize Agent AI based by the types of agent actions and their environments, as illustrated in Fig.~\ref{fig:agentaxes}. Therefore, Agent AI can be broadly grouped into four categories. This section reviews related research~\cite{durante2024agent} and organizes them according to these categories. We also expand on systems combining both intention and manipulation agents in Appendix~\ref{app:Intention}.%, elucidating the Agent AI aspects they capture. We emphasize the significance of integrating VLMs/LLMs and clarify the research domains encompassed by Agent AI. 

\subsubsection{Manipulation action in physical environments}
Agents in this category are intended to work in the physical world, with robotics applications being the typical example~\cite{saycan2022arxiv, huang2022inner, codeaspolicies2022, driess2023palm,brohan2023rt}. Training agents for physical manipulation in an end-to-end manner is typically challenging due to the significant costs associated with collecting a large amount of data for training. Consequently, recent trends have shifted towards solving higher-order task plans with large foundation model and integrating these with lower-level controllers that are trained using conventional methods RT-1~\cite{brohan2022rt} and RT-2~\cite{brohan2023rt} and Agent foundation model~\cite{durante2024foundation}.

\subsubsection{Manipulation action in virtual environments}
This type of agent utilizes a virtual simulated environment. In the robotics domain, the main objective is to train Agent AI through trial-and-error for tasks where physical trials are impractical or risky, including the ability to predict user actions and devise plans for tasks within specific constraints ~\cite{ahn2022can,brohan2023rt,durante2024foundation,gong2023mindagent}. In the case of gaming agents, the goal is not to eventually transition to the physical world, but the learning within the simulation environment itself is the main objective~\cite{park2023generative, wang2023voyager,wang2023describe,baker2022video}.

There have also been a number of works that demonstrate the ability of general-purpose visually-aligned large language models trained on large-scale text, image, and video data to serve as a foundation for creating multi-modal agents that are embodied and can act in various environments \cite{baker2022video,driess2023palm,brohan2023rt, durante2024foundation}. Typically, research on these agents involves simulation platforms for object recognition~\cite{kolve2017ai2, wang2023robogen, mees2022calvin, yang2023octopus, ehsani2021manipulathor, szot2021habitat, puig2018virtualhome, carroll2019utility, li2021igibson, srivastava2022behavior, mittal2023orbit, zhong2023assist, liu2021role, saito2023constraintaware,pmlr-v162-huang22a}.

\subsubsection{Intentional action in physical environment}
A typical example of interactive agents in this category is found in the healthcare domain, such as applications in diagnostics and knowledge retrieval~\cite{lee2023benefits, peng2023check}. In a similar context, several works have developed empathy-aware agents for engaging dialogue and human-machine interactions~\cite{chen2021nice, mao2022biases, wake2023bias, savva2019habitat, puig2023habitat,huang2018turbo}. In other cases, Agent AI's focus on knowledge and logical reasoning involves integrating implicit and explicit knowledge sources. This integration enables more accurate and contextually appropriate responses~\cite{brown2020language, gpt4, lewis2020retrieval, peng2023check, gao2022neural, marcus2019rebooting, gao2020robust, wang2023logical, chen2020mapping, park2023multimodal, li2023camel}.

\subsubsection{Intentional action in virtual environment}
Studies on Agent AI in this category have highlighted the utility for the creation of interactive content in gaming and both VR and XR~\cite{chen2021nice, mao2022biases,huang2023ark}. %In some cases, agents are used to solve abstract concept, such as symbol manipulation~\cite{chen2020mapping,park2023multimodal}. 
Agent navigation following instruacion is also a representative task that falls in this category~\cite{tsoi2022sean, deitke2020robothor}. Similar to gaming agents for intentional action, this type of Agent AI has shown super-human performance in specific games~\cite{meta2022human, yao2023react}.
Recent robotics research also leverages LLMs to perform task planning \citep{saycan2022arxiv, huang2022inner, codeaspolicies2022} by decomposing natural language instruction into a sequence of subtasks, either in the natural language form or in Python code,  then using a low-level controller to execute these subtasks.

\subsection{Multimodel Agent Categorization (Non-Embodied)}
These categories of Agents emphasize the importance of using multimodal information to take beneficial non-embodied from their respective aspects. This indicates the necessity for agents to possess high recognition capabilities for both language and vision, thereby strongly suggesting the effectiveness of leveraging large fondation models. MUltimodel Agent have shown significant utility across a variety of tasks. The advancements in large-scale foundational models and interactive artificial intelligence have opened up novel capabilities for multimodel agent.
A number of works leverage multi-model agents to perform task planning \citep{pmlr-v162-huang22a, wang2023voyager, yao2023react, li2023camel}, and leverage the large multimodels' large internet-scale domain knowledge and zero-shot planning abilities to perform agentic tasks like planning and reasoning. 
Additionally, \cite{huang2022inner}, \cite{codeaspolicies2022}, and \cite{wang2023describe} also incorporate environmental feedback to improve task performance.  

Nevertheless, for agent AI to be genuinely beneficial, they must offer intuitive interaction experiences and adapt to a wide array of environments, contexts, and modalities. To promote research in this area, we proposed a broad range of categorization relevant for multimodal agents without embodied action including~\cite{gui2022vlc,park2023localized}, but not limited to Simulation and environments agents~\cite{puig2018virtualhome}), generative agents~\cite{huang2023ark}, knowledge and logical inference agents ~\cite{lewis2020retrieval, peng2023check, wang2023logical,gui2022kat}, emotion agent ~\cite{chen2021nice}, Neuro-symbolic agents~\cite{chen2020mapping}, and agents for traditional multimodal tasks, multimodal agent systems and infrastructure, and applications of multimodal agents. 

%% file: sections/3_agentAI_applications.tex
\section{Agent AI Application Tasks}
\label{sec:tasks}
In Section~\ref{sec:category}, we categorized existing research within the realm of Agent AI. To offer a tangible understanding of its applications, we introduce four mission-critical domains where Agent AI can have a major impact. 
%[Jianfeng: Sec. 5.4, "interactive multimodality" is NOT a task. probably you can name it "multi-modal content generation"]
%we focus on robotics, gaming, and healthcare. These fields exemplify the diverse potential of Agent AI, aiming to provide readers with a vivid picture of its practical implementations.

%%%%%%%%fig.GPT4vgamediog
%\begin{figure*}
%    \centering
    %\vspace{-1mm}
%    \includegraphics[width=0.98\linewidth]{icml2024/figures/qiuyuan_GPT4vgamediog.png}
%    \vspace{-0.5mm}
%    \caption{\normalsize 
%    The embodied agent for user interactive gaming action prediction and interactive editing with Minecraft Dungeons gaming sense simulation and generation via GPT-4V.
%}
%    \label{fig: GPT4Vgamediog}
%    \vspace{-1mm} 
%\end{figure*}

%%%%%%%%%%Robotics
\subsection{Robotics}
Robots are representative agents that necessitate effective interaction with their environment. In this section, we introduce key elements essential for efficient robotic operation, review research topics where the latest large foundation models have been applied, and share insights from recent studies.

\paragraph{Multimodal Systems}
Recent research focuses on developing end-to-end systems incorporating large foundation model technologies as encoders for input information, guiding robotic actions based on linguistic instructions and visual cues~\cite{huang2018turbo,jiang2022vima,brohan2023rt,brohan2022rt,li2023vision,ahn2022can,shah2023mutex,li2023mastering}.

\paragraph{Task Planning and Skill Training}
Advanced language processing abilities of LLMs interpret instructions and decompose them into robot action steps, advancing task planning technologies~\cite{ni2023grid, li2023interactive, parakh2023human,wake_chatgpt}. For skill training, large foundation models are used for designing reward functions~\cite{yu2023language, katara2023gen2sim, ma2023eureka}, generating data for policy learning~\cite{kumar2023words, du2023video}, or as part of a reward function~\cite{sontakke2023roboclip}.

\paragraph{On-site Optimization}
This involves dynamically adapting and refining robotic skills by integrating task plans with real-time environmental data~\cite{ahn2022can,zhou2023generalizable,raman2023cape,chen2021nice}. Strategies seek to achieve environment-grounded robot execution by adjusting the robot's actions at the task plan or controller level.

\paragraph{Conversation Agents}
LLMs contribute to natural, context-sensitive interactions with humans in conversational robots~\cite{ye2023improved, wake2023gpt}. They process and generate responses that mimic human conversation and estimate conceptual~\cite{hensel2023large,teshima2022deep} and emotional attributes~\cite{zhao2023chatgpt, yang2023evaluations, wake2023bias} of utterances.

\paragraph{Navigation Agents}
Robot navigation focuses on core aspects such as map-based path planning and SLAM~\cite{guimaraes2016ros}. Recent work enables robots to navigate in challenging environments using object names~\cite{chaplot2020object,batra2020objectnav,gervet2023navigating,ramakrishnan2022poni,zhang2021hierarchical} or zero-shot object navigation~\cite{gadre2023cows,dorbala2023can,cai2023bridging}. Vision-Language Navigation (VLN) interprets sentences for navigation in unseen environments~\cite{anderson2018vision,shah2023lm,zhou2023navgpt,dorbala2022clip,liang2023mo,huang2023visual}. VLN interprets sentences rather than object names,
it requires a higher functionality to parse input text~\cite{wang2019reinforced}.

\subsection{Gaming} 
Games provide a unique sandbox to test the agentic behavior of large foundation models, pushing the boundaries of their collaborative and decision-making abilities. We describe three areas in particular that highlight agent's abilities to interact with human players and other agents, as well as their ability to take meaningful actions within an environment.

\paragraph{NPC Behavior}
In modern gaming systems, the behavior of Non-Player Characters (NPCs) is predominantly dictated by predefined scripts crafted by developers. These scripts encompass a range of reactions and interactions based on various triggers or player actions within the gaming environment. 
In light of this situation, Agent AI is at the forefront of revolutionizing NPC technologies. By leveraging large foundation model, Agent AI can provide dynamic dialogues and refine behaviors based on player feedback and in-game data, significantly contributing to the evolution of NPC behavior in games.
%\vspace{-2mm}

\paragraph{Human-NPC Interaction}
Agent AI plays a critical role in enhancing the interaction between human players and NPCs, offering a more immersive gaming experience. The conventional interaction paradigm is primarily one-dimensional, with NPCs reacting in a preset manner to player inputs. Agent AI, utilizing large foundation models, can analyze and learn from human behavior, providing more human-like interactions and increasing realism and immersion~\cite{gong2023mindagent}.
%\vspace{-2mm}

\paragraph{Agent-based Analysis of Gaming}
Gaming is an integral part of daily life, estimated to engage half of the world’s population~\cite{dfcint2020videogameaudience} and exhibits a positive impact on mental health~\cite{granic2014benefits}. Contemporary game systems, however, often exhibit deficiencies in interactions with human players due to primarily hand-crafted behaviors by game developers. 
In such a context, Agent AI proves valuable as a system that analyzes in-game text data, such as chat logs and player feedback, to identify patterns of player behavior and preferences, as well as analyzes image and video data from gaming sessions to understand user intent and actions.

%\vspace{-3mm}
\paragraph{Scene Synthesis for Gaming}
Scene synthesis is essential for creating and enhancing immersive gaming environments, encompassing the generation of three-dimensional (3D) scenes, terrain creation, object placement, realistic lighting, and dynamic weather systems~\cite{huang2023ark}. In modern games, providing vast open-world environments necessitates the use of procedural or AI-driven techniques for automated terrain generation. Agent AI, utilizing large foundation models, aids scene designers by formulating non-repeating, unique landscape design rules based on the designers' desires and the current scene, ensuring semantic consistency and variability of the generated assets. These models expedite object placement and assist in content generation, enhancing the design process.

%%%%%%%%%%%Healthcare
\subsection{Interactive Healthcare}
In healthcare, Agent AI can help both patients and physicians by utilizing large foundation models in understanding the intent of the user, retrieving clinical knowledge, and grasping the undergoing human-to-human interaction, but not limited to these areas. Examples of application include: % However, they come with unique challenges and responsibilities. 

\paragraph{Diagnostic Agents}
LLMs as medical chatbots for patient diagnosis have gained attention for their potential to help triage and diagnose patients, providing equitable healthcare access to diverse populations \cite{lee2023benefits}. They offer a pathway to improve healthcare for millions, understanding various languages, cultures, and health conditions, with initial results showing promise using healthcare-knowledgeable LLMs trained on large-scale web data \cite{durante2024foundation,durante2024agent}. However, risks such as hallucination within medical contexts are notable challenges.

\paragraph{Knowledge Retrieval Agents}
In the medical context, model hallucinations can be dangerous, potentially leading to serious patient harm or death. Approaches using agents for reliable knowledge retrieval \cite{peng2023check} or retrieval-based text generation \cite{guu2020retrieval} are promising. Pairing diagnostic agents with medical knowledge retrieval agents can reduce hallucinations and improve response quality and preciseness.

\paragraph{Telemedicine and Remote Monitoring}
Agent-based AI in Telemedicine and Remote Monitoring can enhance healthcare access, improve communication between healthcare providers and patients, and increase the efficiency of doctor-patient interactions \cite{amjad2023review}. Agents can assist in triaging messages from doctors, patients, and healthcare providers, highlighting important communications, and revolutionizing remote healthcare and digital health industries.

\subsection{Interactive Multimodal Tasks}
%[Jianfeng: "interactive multimodality" is NOT a task. probably you can name it "multi-modal content generation"]

The integration of visual and linguistic understanding is a fundamental of Agent AI. Therefore, the development of Agent AI is closely linked to the performance of multimodal tasks, including image captioning, visual question answering, video language generation, and video understanding. Here are some tasks that have recently garnered significant interest:

\paragraph{Image and Language Understanding and Generation}
Image-language understanding is a task that involves the interpretation of visual content in a given image with language and the generation of associated linguistic descriptions. This task is critical to the development of AI agents that can interact with the world in a more human-like manner. Some of most popular ones are image captioning \cite{mscoco, conceptual-caption, flckr30, krishnavisualgenome}, referring expression \cite{yu2016modeling,karpathy2014deep}, and visual question answering \cite{antol2015vqa,ren2015exploring,singh2019towards}. This demands capabilities beyond object recognition, encompassing a deep understanding of spatial relationships, visual semantics, and integrating world knowledge for accurate descriptive and reasoning abilities.

% More recently, knowledge-intensive Visual Question Answering tasks have been introduced. Multimodal agents should capable of identifying objects in an image, comprehending their spatial relationships, generating accurate descriptive sentences about the scene, and utilizing reasoning skills to handle knowledge-intensive visual reasoning. This requires not just object recognition capabilities, but also a deep understanding of spatial relationships, visual semantics, and the ability to map these visual elements to linguistic constructs with integration of the world knowledge.

\paragraph{Video-Language Understanding and Generation} Video captioning and storytelling involve generating coherent sentences for video frames, challenging due to the need for a comprehensive understanding of each frame and their interrelations. Recent advances leverage large foundation models for improved video-language generation, emphasizing the development of agent-aware text synthesis models for encoding sequences and generating cohesive paragraphs. Video understanding broadens image understanding to include dynamic content and requires agents to interact with visual, textual, and audio modalities. Key tasks include captioning, question answering, and activity recognition, focusing on temporal alignment, sequence handling, and complex activity interpretation. Agents also need to process audio cues like spoken words and background sounds to grasp a video's mood and nuances.
%Our workshop aims to delve into these leader-board, exploring the potential of AI agents to comprehend and interact with dynamic visual content.

% \paragraph{} The creation of agents adept at combining knowledge from diverse sources like visual content, text, and audio is a significant challenge. This knowledge can come from pre-existing databases (explicit knowledge) or learned from the models in their training stage (implicit knowledge). An agent might use its knowledge of common objects and their typical interactions to make sense the visual content with objects and actions described in text. World knowledge can also help an agent understand the significance of certain sounds. Similarly, using world knowledge, an agent can infer the implications of specific sounds, such as a car engine representing an outdoor scene or laughter implying humor. 
Parallel research explores generating scaled datasets from large models, then applying visual instruction tuning \cite{durante2024foundation,durante2024agent, li2023blip, zhu2023minigpt4} on the generated data. Considerable audio, speech, and visual expert perception models are subsequently used to verbalize videos. Speech is transcribed with automatic speech recognition tools, and video descriptions and related data are produced with various tagging, grounding, and captioning models \cite{2023videochat, maaz2023videochatgpt, chen2023videollm, wang2023internvid}. These techniques demonstrate how instruction tuning video-language models on generated datasets may lead to enhanced video-reasoning and communication abilities.

Such agents would be able to understand the context of the video, identify the key steps, and generate a coherent summary of the procedure. This would not only enhance the interpretability of the model but also enable it to provide useful feedback or guidance to the user. 
% To guide the research in this direction, we introduce VideoAnalytica challenge in this workshop, a new dataset and benchmark for analytical video demonstration comprehension that requires sophisticated interaction with the audio-visual-language modalities. We refer to Section \ref{sec:analytica} more details about the challenge.

%\subsection{NLP Task}
%The challenge in interactive AI and NLP has been recognizing task directives and taking action. With advances in deep learning, there's growing interest in studying these areas jointly to improve human-agent collaboration. Key directions for improving language-grounded agents include:
%1)Tool use and querying from knowledge bases: Integrating external knowledge sources like databases and web search into AI reasoning processes. 2) Improved agent reasoning and planning: Enhancing agents' ability to understand complex instructions and predict future scenarios, as demonstrated in ReAct \cite{yao2023react} and other models \cite{yao2023tree}.3)Incorporating system and human feedback: Adapting learning mechanisms to refine strategies and rectify mistakes, exemplified by AutoGen \cite{wu2023autogen}.

We expand upon more cross-modality and  Mix-reality topic discussion in Appendix~\ref{app:Crossmodal}, Appendix~\ref{app:Crossdomain} and~\ref{app:ModalReality}. 

%% file: sections/4_challenge.tex
\section{Deploying Agent AI}
\label{sec:discussion}
%[Jianfeng: the content does not seem to match the title. The issues related to deploying agents include cost-efficiency, data privacy, security, how to leverage user feedback to contiuously improve the system etc.]

We believe that in order to develop a system that incorporates these elements, it is necessary to involve a wide range of experts and practitioners. For instance, there are the following important research areas:

\paragraph{Exploring new paradigms} The development of agent paradigms with integrated modalities (audio, image, text, sensor inputs) may address common issues in large-scale models, such as hallucinations and biases in their outputs, which will enhance their recognition and response capabilities for a wide variety of applications.

\paragraph{General-purpose end-to-end systems} Versatile and adaptable AI solutions can be driven by the development of end-to-end models that are trained with large-scale data.

\paragraph{Methodologies for grounding modalities} By integrating information across various modalities, we can enhance the coherence and efficacy of data processing. We expand on this topic in Appendix~\ref{app:Crossmodal}.

\paragraph{Intuitive human interface} Developing intuitive human interfaces can facilitate effective and meaningful interactions between humans and agents.

\paragraph{Taming LLM/VLMs} Exploring new approaches can address common issues in large-scale foundation models, such as hallucinations and biases in their outputs.

\paragraph{Bridging the gap between simulation and real} 
The "sim-to-real" problem highlights the challenge of deploying AI agents trained in simulations to the real world, where discrepancies in conditions like disturbances and physical properties can degrade performance. To tackle these issues, strategies include:
\begin{itemize}
\item \textbf{Domain randomization} Introducing variability in the simulated environment to better prepare the model for real-world unpredictability~\cite{tobin2017domain,saito2022task}.

\item \textbf{Domain adaptation} Bridging sim-to-real gap by training on both simulated and real-world data~\cite{zhu2017unpaired,rao2020rl,ho2021retinagan}.

\item \textbf{Improvement of simulation} Enhancing simulation fidelity through better replication of real-world conditions~\cite{zhu2017fast,allevato2020tunenet,martinez2020unrealrox,muller2018sim4cv,shah2018airsim,sasabuchi2023task}.
\end{itemize}

\paragraph{Multi-Agent}
Agent AI interaction is currently still a complex process that requires a combination of multiple skills. The current human-machine interaction systems inside multi-agents are primarily effectiveness of cooperation rule-based. They do have intelligent behaviors in response to human/user actions and possess web knowledge to some extent~\cite{gong2023mindagent}. The kind multi agents interactions are very important in the agent development to enable specific behaviors in the agent system design.

\paragraph{Agent Infrastructure and System}
Agent-based AI  is a large and fast-growing community within the domains of entertainment, research, and industry. The development of large foundation models has significantly improved the performance of agent AI systems. However, creating agents in this vein is limited by the increasing effort necessary to create high-quality datasets and overall cost. In industry, building high-quality agent infrastructure has significantly impacted multi-modal agent copilots by using advanced hardware, diverse data sources, and powerful software libraries~\cite{gong2023mindagent}.
%We also introduce ``Agent Infrastructure and System" deployment for your reference in Appendix~\ref{app:infra}.
The rising prevalence of Agent AI underscores the need for robust infrastructure to facilitate their training, evaluation, and deployment. In response to this need, we are introducing a dedicated track for agent research focusing on the infrastructure and methodologies pertinent to the development, evaluation, and deployment of Agent AI. We expect this track will attract a significant number of submissions centered on the efficiency and optimization of agent systems. Agent AI infrastructure is intended to ensure that the broader community can readily access and benefit from these contributions, thereby fostering further advancements in the field.

 We expand on biases and hallucinations in Appendix~\ref{app:Bias} and~\ref{app:Hallucinations} respectively.

%%%%%%%%%%%%%%%%
\section{Challenges for Agent AI}
\label{sec:challenges}
In this paper, we put special emphasis on discovering the current agent AI limitation, and we discuss the challenges ahead for advancing towards deeper and more comprehensive versions of AGI, including the possible need for pursuing a new paradigm that moves beyond next-word prediction.
%[Jianfeng: Need to clarify "new paradigm" -- I think the new paradigm is LLM-powered agents.]

%[Jianfeng: top challenges IMHO include (1) how to augment LLMs trained in static settings to interact w/ dynamic enviroment, e.g., how to cope with when things do not go as planned; (2) how to equip the agents with the abilities of self-improving and continal learning; (3) how to evaluate LLM agents using utility, sociability (e.g., communication efficiency), values (e.g., honesty, harmlessness), and the ability to evolve continually (e.g., continual learning, autotelic learning, adaptability & generalization to new environment).]

Achievement of the Agent AI still have some challenges, especially considering the dynamic system with high modality observations in the physical world. There still exist a number of challenges that need to be addressed, including but not limited to: 1) unstructured environments, where current visual inputs affect both high-level intents and low-level actions of the embodied agent given the same goal instruction; 2) empathy for agent, when open sets of objects, which require the agent's decision-making module to use common sense knowledge that is hard to encode manually; 3) multi-agent interactions and collaborations, which require the agent to understand and operate on more than just template-based commands, but also a context of goals, constraints, and partial plans expressed in everyday language. To enable a more comprehensive approach to these complex challenges, the inclusion of researchers and practitioners from a broader range of fields is critical. 4) Emergent ability for embodied large agent foundation model. %Please refer to Appendix~\ref{app:EA} and find more details.
We aspire to broaden our collective understanding of the potential and limitations of Agent Paradigm by leveraging our unique and diverse perspectives. We strongly believe that this proposed new agent paradigm will not only enrich the perspectives of individual practitioners, but will also enhance the community's collective knowledge and promote a holistic view that is more inclusive of the wide-ranging challenges faced by future agent AI.

\section{Emergent Abilities}
\label{app:EA}
Despite the growing adoption of interactive agent AI systems, the majority of proposed methods still face a challenge in terms of their generalization performance in unseen environments or scenarios. 
Current modeling practices require developers to prepare large datasets for each domain to finetune/pretrain models; however, this process is costly and even impossible if the domain is new. To address this issue, we propose building interactive agents that leverage the knowledge-memory of general-purpose foundation models (ChatGPT, Dall-E, GPT-4, etc.) for novel scenarios, specifically for generating a collaborative space between humans and agents. We discover an emergent mechanism–- which we name Mixed Reality with Knowledge Inference Interaction–-that facilitates collaboration with humans to solve challenging tasks in complex real-world environments and enables the exploration of unseen environments for adaptation to virtual reality. For this mechanism, the agent learns i) micro-reactions in cross-modality: collecting relevant individual knowledge for each interaction task (e.g., understanding unseen scenes) from the explicit web source and by implicitly inferring from the output of pretrained models; ii) macro-behavior in reality-agnostic: improving interactive dimensions and patterns in language and multi-modality domains, and make changes based on characterized roles, certain target variable, influenced diversification of collaborative information in mixed-reality and LLMs. We investigate the task of knowledge-guided interactive synergistic effects to collaborated scene generation with combining various OpenAI models, and show promising results of how the interactive agent system can further boost the large foundation models in our setting. It integrates and improves the depth of generalization, conscious and interpretability of a complex adaptive AI systems.

%% file: sections/5_influences.tex
\section{Impact Statement}
\label{app:broaderimpact}

 Agent AI paradigm is to create general-purpose agents that can work alongside humans in both real and virtual environments. This paradigm therefore intends to have a very broad impact, possibly affecting all members of society.
Agent AI framework emphasizes the integration of agents into the wider environment across a variety of settings, such as gaming, robotics, healthcare, and long-video understanding. Specifically, the development of multimodal agents in gaming could lead to more immersive and personalized gaming experiences, thereby transforming the gaming industry. In robotics, the development of adaptive systems could revolutionize industries ranging from manufacturing to agriculture, potentially addressing labor shortages and improving efficiency. In healthcare, the use of large foundation model as diagnostic agents or patient care assistants could lead to more accurate diagnoses, improved patient care, and increased accessibility to medical services, particularly in underserved areas. Furthermore, the ability of these models to interpret long-form videos could have far-reaching applications, from enhancing online learning to improving technical support services. In general, the Agent AI framework will have significant downstream effects on a wide range of industries and people across the world.

We must also highlight the diverse and complex challenges that come with implementing AI agents across a wide variety of environments and situations. For instance, there are many limitations and potential hazards linked to Agentic AI systems when they are developed for specialized sectors such as healthcare diagnostics. In this domain, issues like dangerous hallucinations in AI behavior can pose significant risks, highlighting the critical need for meticulous design and testing. However, these specific challenges may not be equally relevant or noticeable when considering AI agents crafted for the gaming industry. In such recreational fields, developers might instead prioritize tackling different hurdles, such as the need for AI to perform more open-ended generation and exhibit creativity, adapting dynamically to unpredictable gameplay scenarios and player interactions.

%% file: sections/6_conclusion.tex
\section{Conclusion}
\label{sec:conclusion}
Our proposed Agent AI focuses on advanced multimodal systems that interact effectively within both physical and virtual environments and facilitate effective interaction with humans. %This paper will bring together researchers in the field of agent AI with expertise in large foundation model based embodied modules in exploring the holistic intersections. By leveraging the collective expertise of agent paradigm, agent foundation model, agent infrastructure, and agent system from various AI disciplines. 
This paper aims to unite researchers to deepen the discourse on Agent AI, cutting across various AI disciplines including agent paradigms, foundation models, infrastructures, and systems. 
%This paper aims to not only advance scientific interactive understanding of Agent AI, but also to discuss the embodied agent at the frontier of novel holistic intelligence research and helps us position ourselves to capitalize on emerging foundational models. 
Our goal is to enrich the scientific comprehension of Agent AI and explore the potential of embodied agents within the realm of holistic intelligence research. This endeavor positions us to leverage emerging foundational models effectively.

%As such, we intend to discuss important topics, including but not limited to: 1) Application of foundation models: the development of agents with integrated modalities (audio, image, text, sensor inputs), aiming to enhance their recognition and response capabilities for a wide variety of applications.
%2) General-purpose end-to-end systems: the development of end-to-end models that are trained with large-scale data, seeking to create versatile and adaptable AI solutions.
%3) Methodologies for grounding modalities: integrating information across various modalities, enhancing the coherence and efficacy of data processing.
%4) Intuitive human interface: the development of effective and meaningful interaction between humans and agents.
%5) Taming LLM/VLMs: exploring new approaches to address common issues in large-scale models, such as hallucinations and biases in their outputs. 

%We aspire to broaden our collective understanding of the  potential and limitations of agentic AI by leveraging our unique and diverse perspectives. We believe that this approach will  enhance the community's collective knowledge and promote a holistic view that is more inclusive of the wide-ranging challenges faced by future AI.

\clearpage

%% file: sections/7_broade_impact.tex
\section*{Ethical Consideration} 
\label{sec:ethics}

Agent AI systems have many applications. In addition to interactive AI, grounded multimodal models could help in generating training datasets for robots and AI agents, and assist in productivity applications, helping to re-play or paraphrase scenario, predict actions in novel scenarios, or synthesize 3D or 2D scenes. Fundamental advances in agent AI help contribute towards these goals and many would benefit from a greater understanding of how to model embodied and empathetic behavior in a simulated environment or the real world. Therefore, there are many applications that have positive benefits.

However, this technology could also be used by bad actors. Agent AI systems that generate content can be used to manipulate or deceive people. Therefore, it is very important that this technology is developed in accordance with responsible AI guidelines. For example, explicitly communicating to users that content is generated by an AI system and providing the user with controls in order to customize such a system. It is possible the Agent AI could be used to develop new methods to detect manipulative content - partly because it is rich with hallucinations that emerge from large foundation models - and thus help address another real world problem.

For example, ethical deployment of large agents foundation models, especially in sensitive domains like healthcare, is paramount. AI agents trained on biased data could potentially worsen health disparities by providing inaccurate diagnoses for underrepresented groups. Moreover, the handling of sensitive patient data by AI agents raises significant privacy and confidentiality concerns. In the gaming industry, AI agents could transform the role of developers, shifting their focus from scripting non-player characters to refining agent learning processes. Similarly, adaptive robotic systems could redefine manufacturing roles, necessitating new skill sets rather than replacing human workers. Navigating these transitions responsibly is vital to minimize potential socio-economic disruptions.

Furthermore, the agent AI focuses on learning collaborative policies in simulation and there is some risk of directly applying the policy to the real world due to the distribution shift. Robust testing and continuous safety monitoring mechanisms should be put in place to minimize risks of unpredictable behaviors in real-world scenarios.

%We further analyze broader impacts of Agent AI research in Appendix~\ref{app:broaderimpact}.

\section*{Limitations}
\label{sec:limitations}

The main thesis of our work is that the Agent AI formulation helps to bring the field of AI back to its roots in holistic intelligence. However, there are still many unknowns within the Agent AI paradigm. Existing foundation models exhibit biases and hallucinations, and it is unclear whether these can be resolved through scaling up model and dataset sizes or if these are fundamental limitations of Agent AI.

We also acknowledge that there are many additional challenges in this field that we have not covered in Section~\ref{sec:challenges}. As a growing field with a potential for major impact, we believe that the development of Agent AI must include a diverse range of perspectives across disciplines to ensure that it has a positive impact on humanity.